\documentclass[letterpaper, 10 pt, conference]{ieeeconf}
\IEEEoverridecommandlockouts
\usepackage{amsmath} % assumes amsmath package installed
\usepackage{amssymb}  % assumes amsmath package installed
\usepackage{hyperref}
\usepackage{placeins}
\usepackage{array}
\let\labelindent\relax
\usepackage{enumitem}
\usepackage{url}
\usepackage{stackengine}
\newcolumntype{P}[1]{>{\centering\arraybackslash}p{#1}}
\usepackage{multirow}
\usepackage{comment}
\newcommand{\specialcell}[2][c]{%
  \begin{tabular}[#1]{@{}c@{}}#2\end{tabular}}
\usepackage{ctable} % for \specialrule command

\title{\LARGE \bf
Let's move on: Topic Change in Robot-Facilitated Group Discussions
}

\author{Georgios Hadjiantonis$^{1}$, Sarah Gillet$^{1}$, Marynel {V\'{a}zquez}$^{2}$, Iolanda Leite$^{1}$ and Fethiye Irmak Dogan$^{1}$% <-this % stops a space
\thanks{$^{1}$KTH Royal Institute of Technology, Sweden, {Contact: \tt\small \{ghad, sgillet, iolanda, fidogan\}@kth.se}. This work was supported by the S-FACTOR project from NordForsk, the Swedish Foundation for Strategic Research (SSF FFL18-0199), and the Wallenberg Al, Autonomous Systems and Software Program (WASP) funded by the Knut and Alice Wallenberg Foundation.}
\thanks{$^{2}$Yale University, USA,
        {Contact: \tt\small marynel.vazquez@yale.edu}. Marynel was supported by the National Science Foundation (IIS-2143109).}%
}

\begin{document}

\maketitle
\thispagestyle{empty}
\pagestyle{empty}

%%%%%%%%%%%%%%%%%%%%%%%%%%%%%%%%%%%%%%%%%%%%%%%%%%%%%%%%%%%%%%%%%%%%%%%%%%%%%%%%

\begin{abstract}
Robot-moderated group discussions have the potential to facilitate engaging and productive interactions among human participants. Previous work on topic management in conversational agents has predominantly focused on human engagement and topic personalization, with the agent having an active role in the discussion. Also, studies have shown the usefulness of including robots in groups, yet further exploration is still needed for robots to learn when to change the topic while facilitating discussions. Accordingly, our work investigates the suitability of machine-learning models and audiovisual non-verbal features in predicting appropriate topic changes. We utilized interactions between a robot moderator and human participants, which we annotated and used for extracting acoustic and body language-related features. We provide a detailed analysis of the performance of machine learning approaches using sequential and non-sequential data with different sets of features. The results indicate promising performance in classifying inappropriate topic changes, outperforming rule-based approaches. Additionally, acoustic features exhibited comparable performance and robustness compared to the complete set of multimodal features. Our annotated data is publicly available at \href{https://github.com/ghadj/topic-change-robot-discussions-data-2024}{https://github.com/ghadj/topic-change-robot-discussions-data-2024}.
\end{abstract}

%%%%%%%%%%%%%%%%%%%%%%%%%%%%%%%%%%%%%%%%%%%%%%%%%%%%%%%%%%%%%%%%%%%%%%%%%%%%%%%%

\section{INTRODUCTION}
% TODO one first sentence why discussions in teams are important
Group discussions are commonly used for brainstorming and making informed decisions, promoting collaboration, diversity, and innovative thinking~\cite{forsyth2014group}. However, managing the conversation flow can be challenging for the participants. In such cases, a moderator can play an essential role in ensuring that the discussion is engaging, the participants establish a common ground, and transitions between discussion points are smooth and timely. In this work, we explore how a robot could facilitate a group discussion by deciding \textit{when} a transition between topics is needed and appropriate. 
%Research on the moderation of focus groups has emphasized the need for specific skills and various challenges in coordinating a constructive discussion, on top of managing related practicalities, e.g., operating recording equipment and taking notes \cite{wilkinson_focus_1998}. Due to these challenges, there is often a need for a co-moderator or assistant.

% TODO maybe we should talk about related work on moderating group discussions rather than work on group dynamics, I have a few open tabs on this and will add tomorrow
Recent research in group Human-Robot Interaction (HRI) has highlighted the impact of robot behavior on the quality and effectiveness of the interaction, as well as the ability of the robot to influence the individuals and shape group dynamics \cite{sebo_robots_2020, ISRgillet2024}. Such examples demonstrate that robots can have a positive impact on the level of verbal communication among adults in care facilities \cite{sabanovic_paro_2013, thompson_robot_2017}, balance participation during team decision-making discussions \cite{tennent_micbot_2019}, and improve task performance and group cohesion \cite{Strohkorb2016, short_robot_2017}.
%, and the reflection of robots' behavior in people's interpersonal interactions~\cite{strohkorb_sebo_ripple_2018}. %Given the demonstrated impact robots have on increasing pro-social behavior among group members and influencing group dynamics, there is a potential for social robots to be used as moderators of group discussions among humans.

\begin{figure}
    \centering
     \setstackgap{S}{0.2\baselineskip}
    \Shortstack[l]{\stackinset{l}{ .05in}{t}{.05in}{\textcolor{gray}{}}{\includegraphics[width=0.36\textwidth]{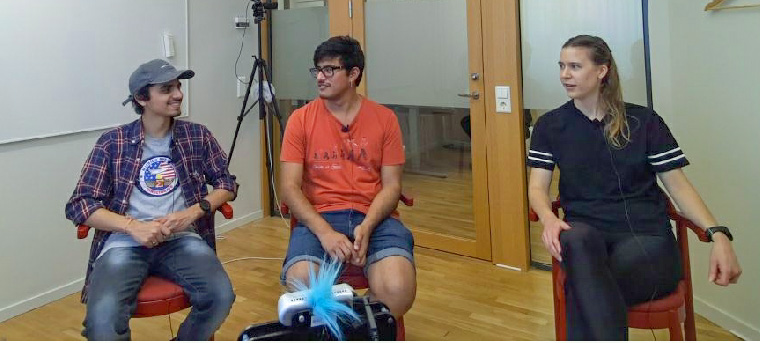}} \stackinset{l}{ .05in}{t}{.05in}{\textcolor{gray}{}}{\includegraphics[width=0.36\textwidth]{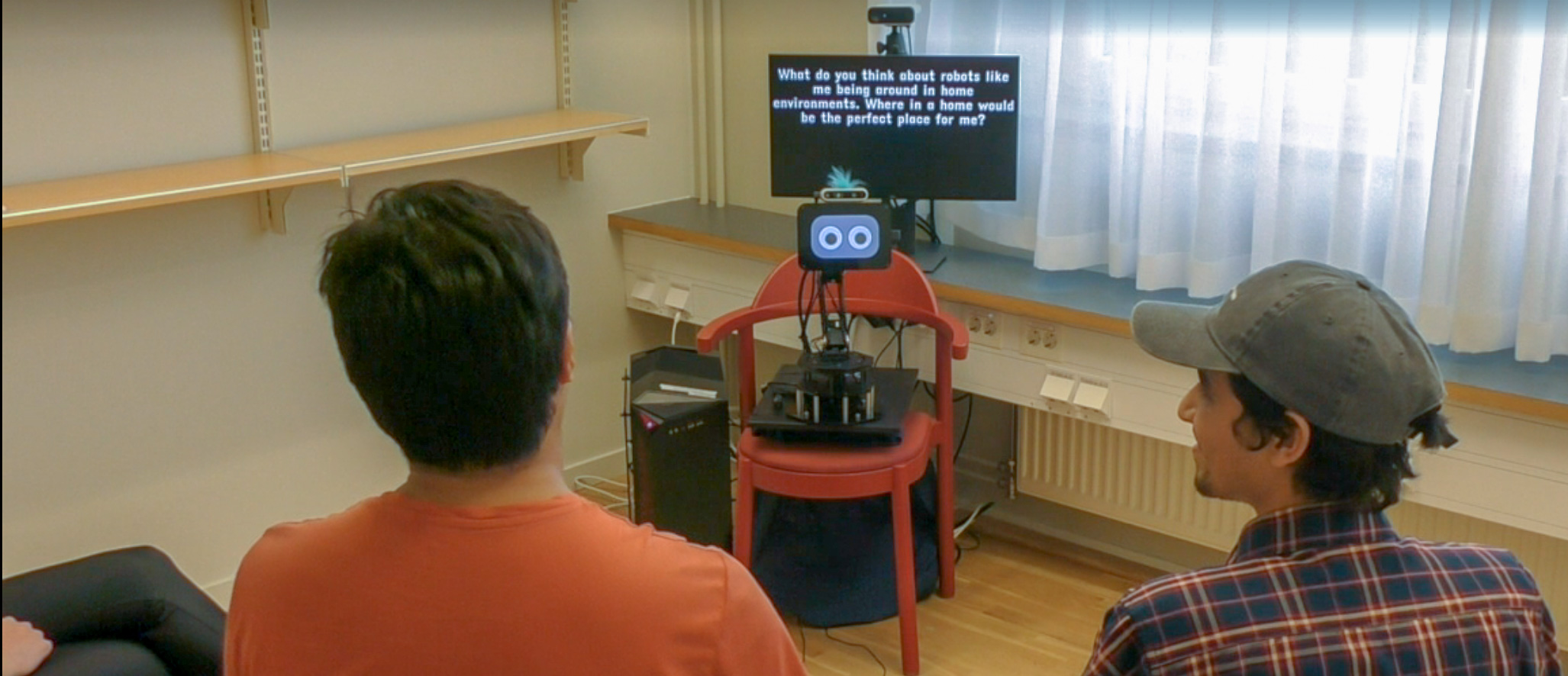}}
    }
    \vspace{-1em}
\caption[Frames of the robot interaction from two perspectives.]{The interaction between the robot and three participants from two different perspectives. The robot moderates the group discussion and needs to decide when to move on to the next topic.}
\vspace{-1em}
\label{fig:frames-setup}
\end{figure}

This work investigates how robot facilitators can autonomously decide when to change a discussion topic.  Unlike prior work, which typically decides when to change topic through a Wizard-of-Oz paradigm \cite{de2023co,10.3389/frobt.2021.657291} or by using handcrafted heuristics \cite{mizrahi2022vrobotator}, our work focuses on endowing robots with the ability to moderate the topic of a discussion between humans. Using a group HRI dataset collected with the setup depicted in Figure~\ref{fig:frames-setup}, we explore the possibility of having robots decide in a fully autonomous manner on topic changes during discussions, as long as the robot does not speak over people. 

%We approach the challenge from the perspective of detecting if a discussion topic ends and the potential speakers do not intend to continue - a topic change is \textbf{appropriate}. We also aim to detect if the group is done discussing the topic, and awkward silence could be a result - a topic change is \textbf{needed}. In other moments, the robot might decide that a topic change is \textbf{not needed}.

%This work provides an initial exploration of possible paths to automating the decisions on topic change and treats 
We frame the problem of deciding on topic changes as a classification task, i.e., the robot decides, given a set of multi-modal features,  if a topic change is \textit{needed}, \textit{appropriate}, or \textit{inappropriate}. We investigate leveraging information beyond verbal cues in this work and evaluate a variety of models in
a content-free manner that is not limited to transcription or speech recognition of the conversations. In particular, our set of features is informed by research concerning content-free approaches to topic segmentation~\cite{kovacs_topical_2016, hunyadi_temporal_2020, tomiyama_identifying_2018} and related linguistic research on cues that describe the structure of discourse topics and turn-taking. We evaluate these multi-modal features on their promise for this problem and study varying machine learning models using a time sequence of feature data (sequential modeling) and aggregated individual feature data (non-sequential modeling) for topic change prediction. 

To our knowledge, this paper is the first to consider the problem of decision-making for changes in a discussion topic in robot-moderated discussions. % and also uses learning-based methods to address this challenge.
In summary, our work makes the following contributions:
\begin{itemize}[leftmargin=*]
    \item[--] We address an unsolved decision-making problem in HRI: when should a robot moderator change the topic of a group discussion among multiple people?
    \item[--] We evaluate multi-modal features and assess their importance to the multiparty topic-change problem.
    \item[--] We investigate how to apply sequential and non-sequential input-based machine learning models to the topic-change problem and evaluate their performance.
    \item[--] We contribute an annotated dataset for benchmarking topic-change prediction algorithms and expanding research on this problem domain. 
\end{itemize}

%As such, the problem of topic change is related to content-free approaches to topic segmentation\cite{kovacs_topical_2016, hunyadi_temporal_2020, tomiyama_identifying_2018}. The difference in our setup is that the robot has only the momentary information of the conversation available and not a completed discussion.
% I am thinking that maybe we just argue that the decision is made two seconds after a turn ends

% Consequently, we build upon related linguistic research on the cues in the structure of discourse topics, turn-taking, and topic segmentation/change detection, to investigate the features and models for such decision-making. Nevertheless, there is an inherent overlap between research on topic structure and conversational engagement.

% Specifically, we aim to answer the following research question about the decision-making system:

% \textit{RQ1: Can machine learning models be used to predict when the topic of the discussion should change? How effective are sequential data modeling methods and non-sequential data modeling methods for this task?}

% In addition, we seek to answer the following research question relevant to the non-verbal cues:

% \textit{RQ2: What audiovisual features could help a robot decide whether to take the initiative and introduce a new topic/question or when to continue probing or waiting for the discussion to finish?}

%Using multi-modal features, i.e. audio signals and body movement, we explore different models to predict if a topic change is \textbf{needed}, \textbf{appropriate}, or \textbf{inappropriate}.
\section{Related work}

% Robots have critical impacts on group dynamics~\cite{sebo_robots_2020}. In groups, they have served as moderators of collaborative games~
% \cite{short_robot_2017, vazquez2015social}, also been used in co-design sessions~\cite{de2023co} and in educational settings~\cite{mizrahi2022vrobotator}. These studies have demonstrated critical findings about the impacts of robots in groups, which include facilitating brainstorming~\cite{10.3389/frobt.2021.657291, vazquez2017towards}, promoting participant engagement and boosting task performance in team discussions~\cite{tennent_micbot_2019}.
The study of robots and groups has gained importance in HRI~\cite{NonDyadicSchneiders22}, 
including how robots shape and facilitate group dynamics~\cite{sebo_robots_2020, ISRgillet2024}. 
In particular, robots have been shown to improve conflict situations  \cite{Martelaro2015, Shen2018}, provide emotional support \cite{erel2021enhancing}, and foster the expression of vulnerability \cite{StrohkorbSebo2018} and group cohesion~\cite{Strohkorb2016}. Further, prior work studied how robots could support the process of inclusion among adults~\cite{StrohkorbSebo2020StrategiesTeams}, and children~\cite{gillet2020mediator, tuncerSmileInclusion}, shape participation behavior \cite{tennent_micbot_2019, RobotLevelGilletCumbal2021}, moderate collaborative games \cite{vazquez2015social, short_robot_2017}, facilitate educational activities \cite{mizrahi2022vrobotator} or brainstorming sessions~\cite{10.3389/frobt.2021.657291, vazquez2017towards}.

In group settings, discussion topics play a critical role in people's involvement in the interactions. The problem of selecting a discussion topic has traditionally been studied with conversational agents which aim to cohesively select appropriate topics~\cite{grassi_knowledge-grounded_2022} or personalize them based on user engagement~\cite{glas_topic_2018}. Other work has focused on how to guide conversations into a target subject~\cite{tang_target-guided_2019} or to bridge different topics~\cite{sevegnani_otters_2021}. While our work addresses aspects of conversational management, it does not involve topic selection. Our work focuses on predicting \textit{when} a topic should change.

Topic change is a critical aspect of ensuring people's engagement in group discussions, which can be derived from verbal cues and non-verbal indicators~\cite{rich_recognizing_2010, hutchison_direction_2005, ishii_gaze_2013}. Recent studies on topic segmentation have suggested multi-modal approaches and acoustic-based methods, using prosodic and visual features~\cite{kovacs_topical_2016, hunyadi_temporal_2020, tomiyama_identifying_2018, eisenstein_gestural_2008}. In parallel, linguistic research has suggested a variety of cues indicating topic boundaries and turn-taking, including prosodic features~\cite{nakajima_study_1993, herman_phonetic_2000, swerts_melodic_1994}, speech rate~\cite{zellers_prosodic_2011}, body gestures~\cite{quek_gestural_2002, duncan_signals_1972, zellers_prosody_2016, auer_previews_1992, cassell_non-verbal_2001}, and gaze~\cite{goodwin_concurrent_1987, sidnell_gaze_2012, quek_gesture_2000}. Our work builds upon the prior understanding of the features that are correlated with the end of discussion topics for determining appropriate topic changes.  Additionally, unlike topic segmentation, our approach does not utilize a fully completed conversation but only uses momentary information from the discussion to proactively make real-time topic change decisions.

\section{Dataset}
We used a dataset collected from interactions between the Shutter robot~\cite{10.1145/3610978.3641090} and groups of human participants (16 groups of two, and nine groups of three participants) as depicted in Figure \ref{fig:frames-setup}. The group was asked to brainstorm about robots in home environments, while the robot had the role of moderating the discussion. Data was collected through individual close-talk microphones and the body tracking module of an Azure Kinect Camera, which was placed behind the robot.

During the discussions, the robot presented various topics and asked leading questions to guide the discussion. For example, the robot would ask: \textit{``What do you think about robots like me being around in home environments?"} Further, the robot could ask follow-up questions to encourage more ideas and deepen the discussion, e.g., \textit{``Do you have other ideas to share?"}. To determine whether to move on to the next topic or ask a follow-up question, the robot used a simple rule-based heuristic devised by an expert: a topic change would be initiated after a total of 60 seconds of speech when none of the participants was speaking (not speaking was based on a 2 seconds silence threshold). We use this heuristic as a baseline method in this work (described in Section~\ref{sec:baselines}).

\vspace{-0.3em}
\subsection{Annotations}
The dataset was manually annotated by the first author in order to identify whether, at the end of each utterance, the robot should change the topic (i.e., the change is \textit{\textbf{needed}}), could change the topic (the change is \textit{\textbf{appropriate}}) or should wait for more contributions (the change is \textit{\textbf{inappropriate}}).
%identify {\color{red}please add here a high-level description of the goal of the annotation}. 
The annotator watched the video recordings from the robot’s point of view and focused on labeling robot decisions two seconds after an utterance ended. This approach ensured that the robot would wait until human participants concluded talking and allowed it to capture valuable insights from the moments directly after one participant's contribution. 

We used voice activity detection with a silence threshold of 750 msec to detect utterances, which is slightly longer than what was previously used in turn-taking prediction (200 and 500 msec \cite{skantze_turn-taking_2021}), due to the less demanding response and to account for ``search" or ``repair pauses", i.e., while a speaker pauses to search for an appropriate word or phrase, or attempt to revise what was previously stated \cite{nakajima_study_1993}. In total, 1529 utterances were extracted from 2-participant sessions and 930 utterances from 3-participant sessions.

 %The annotation determined whether, at the end of each utterance, the robot should change the topic (i.e., the topic change is \textit{\textbf{needed}}), could change the topic (the topic change is \textit{\textbf{appropriate}}), or should wait for more contributions (the topic change is \textit{\textbf{inappropriate}}). %Specifically, each utterance was assigned to one of the three following classes:

% \begin{enumerate}
%     \item \textbf{Needed}: Topic change needed represents cases where the robot should have changed the topic and stopped probing. This includes instances where it was clear that the participants felt frustrated, uncomfortable, or not interested anymore in the current topic and had nothing more to say. 
    
%     \item \textbf{Not appropriate}: Topic change not appropriate represents cases where the robot should have stayed on the current topic and either waited or, when given a chance, probed the participants to talk. For instance, when the participants were still engaged in the conversation, willing to continue talking or thinking, or when their answers were very short and not sufficient. 
    
%     \item \textbf{Appropriate}: Topic change appropriate represents cases where both decisions of changing topic or probing/waiting could apply. For example, given the discussion, the participants could still have something more to add if probed with another question, but the next topic could also be introduced.
% \end{enumerate}

Note that the annotation decision was made considering the whole group interaction,  not just considering only the active participant. Figure~\ref{fig:utterance-sequence} provides an extract of the interaction between two participants and the robot as well as the corresponding annotation per utterance.% (Fig.~\ref{fig:utterance-sequence})  as well as the class distribution (Tables~\ref{tab:number-elements-session-2} and~\ref{tab:number-elements-session-3}). % are attached in the Appendix. 
%( in Fig.~\ref{fig:utterance-sequence}. The tables~\ref{tab:number-elements-session-2} and~\ref{tab:number-elements-session-3} show the distribution of utterances assigned to each class, for sessions with two and three participants, respectively.)

\begin{figure}[t!]
\centering
\footnotesize
\begin{tabular}{|p{0.3cm}p{0.3cm}p{4.5cm}c|}
\hline
   &     &                                                                  & annotation            \\
1  & R:  & is there anything that someone could find concerning?            &       -               \\
2  & P2: & concerning? hmm...                                               & not appropriate       \\
3  & P1: & about the room or about what?                                    & not appropriate       \\
4  & P2: & I think about the robot being in the room                        & not appropriate       \\
5  & P1: & I guess in this type of room there is not a lot of privacy       & not appropriate       \\
6  & P2: & yeah, the living room is the bedroom, so that is kind of tricky  & appropriate           \\
7  & P1: & yeah...                                                          & appropriate           \\
   & \multicolumn{3}{p{6cm}|}{{[}both participants turn towards the robot{]}} \\
8  & R:  & do you have other ideas to share?                                &       -               \\
9  & P2: & hmm...                                                           & not appropriate       \\
10 & P1: & hmm...                                                           & not appropriate       \\
11 & P2: & hmm... no...                                                     & not appropriate       \\
12 & P1: & no... I imagine this type of robot in big houses but not really in student rooms or student apartments. uhmm... & not appropriate \\
13 & P1: & I see the limitation there, probably.                            & needed                \\
   & \multicolumn{3}{p{6cm}|}{{[}both participants turn towards the robot{]}} \\
\hline
\end{tabular}
\caption[Utterance sequence between the robot and two participants.]{Utterance sequence from our annotated dataset. The interaction is between the robot (R) and two participants (P1~and~P2), discussing the question: ``Is there any way that a robot like me could be helpful in those places in your home you just described?".}
\label{fig:utterance-sequence}
\vspace{-1em}
\end{figure}

\vspace{-0.3em}
\subsection{Feature Extraction}
We created a feature vector for topic-change classification using the data collected from group interactions. The feature vector corresponded to each utterance and included acoustic attributes of the current speaker, hand gestures, body and head movements of all the participants, and the total duration of each utterance.
%Regarding the time frame of the features, previous work on turn-taking prediction used acoustic features during the last 200-1000 msec of speech \cite{gravano_turn-taking_2011, johansson_opportunities_2015}. In another work in topic change detection, the authors reported performance improvement when using a more extended context to classify instances, i.e., 2.56 seconds before and after the decision time \cite{kovacs_topical_2016}. 
Previous work on turn-taking prediction considered windows of 200-1000 ms at the end of speech for computing acoustic features \cite{gravano_turn-taking_2011, johansson_opportunities_2015}. Inspired by work in topic change detection, indicating improved performance using an extended context of 2.56 seconds before and after the decision time \cite{kovacs_topical_2016}, we experimentally determined to extract all the features during the time interval from 2 seconds before to 2 seconds after the end of each utterance.

We created a fixed-size feature representation for topic-change classification independent of the group size in the interaction because the dataset contains groups of two and three participants. For acoustic features, we only used the features of the active speaker. For all other features, we used one set of features of the active speaker and a second set as an average over the features of the remaining participants. The features are further detailed below:

\begin{description}[align=left,leftmargin=0em,labelsep=0.2em,font=\textbf,itemsep=0em,parsep=0.3em]
\item[Acoustic Features: ] These features were extracted from the individual audio signals of the active participants. Specifically, we computed the mean, maximum, and minimum value and standard deviation of the speech energy and pitch over the given data window. Additionally, we calculated the mean value of the voice quality features, i.e., jitter, shimmer, and Harmonics-to-Noise Ratio (HNR). Since pitch and voice quality features are characteristics of voice, only the last 2 seconds before the end of the utterance were considered. 

\item[Hand Gestures and Body Features: ] The features were computed from the Kinect body joints. For upper body movements, we calculated the relative position of the chest to the pelvis in the $x$ and $y$ axes, capturing if a participant was leaning forward or sideways, respectively. To capture the hand movements, we calculated the 3D position of the left and right hands, respectively, in relation to the chest. In addition, we included hand velocity and upper body movement by calculating the temporal difference between successive data points. For a detailed definition of the Kinect coordinate system and body tracking joints, refer to the Microsoft Azure Kinect documentation \footnote{\href{https://learn.microsoft.com/en-us/azure/kinect-dk/body-joints}{https://learn.microsoft.com/en-us/azure/kinect-dk/body-joints}}.

\item[Head Rotation Features: ]As a proxy for gaze direction, we computed the relative head rotation between the participants and the robot. First, we measured the relative horizontal head rotation angle between the participants and the robot. To capture the head direction in relation to the interactants independently of their position in the room, the rotation angle between a human and the robot was transformed into one of three values as follows. First, for the human speaker, 0 corresponded to looking at the robot, 1 looking at a listener, and -1 looking away from both the listeners and robot. For a human listener, 0 corresponded to looking at the robot, 1 indicated looking at the speaker, and -1 was looking away in another direction. 
%the rotation angle was scaled between -1 and 1, corresponding to the following cases: a value of 0 in the case the participants were looking directly at the robot;  a value of 1 when fully rotating to the listener/speaker; and a value of -1 when rotating fully at the opposite direction. For a speaker at the center of the group, the scale was symmetric, i.e., increasing to 1 towards left and right participants. 
Similar to the hand and body features, we further calculate the temporal difference of successive head direction values to capture the head rotation movement in addition to head direction.
\end{description}

%\section{METHODOLOGY}

\section{METHODOLOGY}
\subsection{Problem Formulation}
We define the problem of determining when to change the discussion topic at the end of an utterance as a classification problem. The goal is to learn a classification function $f$ that maps input features $x$ to a predicted class label $\hat{y}$: 
\begin{equation}
    f: x \mapsto \hat{y}
    \label{eq:x-y}
\end{equation}
%For the non-sequential modeling approach, $x \in \mathbb{R}^N$, with $N$ the total number of features, aggregated separately for the window of $t$ seconds before the end of each utterance and the window starting from the end until the $t$ seconds that followed; for the sequential modeling approach, $x~=~[x_1, x_2, ..., x_\tau]$, where $\tau$ the length of the sequence between $t$ seconds before and after the end of an utterance, and $x_i \in \mathbb{R}^N$, for $N$ the total number of features at each instance; 
\noindent where the label $\hat{y}  \in \mathcal{Y}$  can take one of three values, $\mathcal{Y} = \{\text{\textit{not appropriate, appropriate, needed}}\}$.
%We propose a non-sequential and a sequential data modeling approach as $x$ and explore different functions $f$ handling this data input. The two types of approaches are further detailed next.
To model the input $x$, we propose non-sequential and sequential data modeling approaches and explore different functions $f$ handling these data inputs. The two types of approaches are detailed next.

\subsection{Non-sequential Data Modeling Approach}

For the non-sequential data modeling approach, each feature is aggregated separately for two-time windows: (1)~from 2 seconds before until the end of each utterance, and (2)~starting from the end of each utterance until 2 seconds afterwards. This results in an input feature vector  $x_i \in \mathbb{R}^N$ for an utterance $i$, where $N$ is the total number of features. %, aggregated before and after the end of the utterance. %Duration $t$ is experimentally set as 2 seconds.

Using the above features, we approximate the function $f$ from eq. (\ref{eq:x-y}) with the following models: a Decision Tree (DT), a Random Forest (RF), a Support Vector Machine (SVM), and a Multilayer Perceptron (MLP) classifier. 

The models were trained using the same dataset format, i.e., for $M$  total utterances, $X = [x_1, ..., x_M]$ and $y = [y_1, ..., y_M]$, where $x_i \in \mathbb{R}^N$ and $y_i \in \mathcal{Y}$. % (eq.~\ref{eq:Y}) the target class label for utterance $i$. 
Given a new input vector $x$, the models output a predicted class label $\hat{y} \in \mathcal{Y}$.

Regarding DTs, we used the Gini impurity measure as a split criterion and tuned the max depth of the tree and the min samples per split. Similarly, for RFs, we used the Gini impurity as the split criterion and tuned the number of estimators and maximum depth of trees. For SVMs, we used the Radial Basis Function (RBF) as the kernel function. The kernel function, gamma, and regularization term were tuned (see Section~\ref{sec:training-and-parameter-optimization} for training and tuning procedure). 

For MLPs, we experimented with one and two hidden layers of different sizes with Rectified Linear Unit (ReLU) activation(s) \cite{nair_rectified_2010}. For the output layer of the network, we used a softmax activation function, estimating the probability distribution over the desired classes. %Thus, given an input feature vector $x$ and model parameters $\theta$ (i.e., weights and biases), the mapping of input vector to the conditional probability can be written as
%\begin{equation}
%    x \rightarrow P(y|x;\theta)
%    \label{eq:x-pyx}
%\end{equation}
%
%The predicted class was the one with the highest probability. 
%To train the model, the target values, $y$, were represented with one-hot encoding vectors.
For model supervision, we used the categorical cross-entropy loss. Model parameters were optimized using the Adaptive moment estimation algorithm (Adam) \cite{kingma_adam_2015}. The batch size used during training was also tuned, considering batch sizes of 8, 16, and 32.

\subsection{Sequential Data Modeling Approach}
In contrast to the non-sequential modeling, sequential data is processed as a series of vectors with temporal structure. The sequential features, as the non-sequential ones, were the same types of features and contained information for the period of 2 seconds before until 2 seconds after the end of the utterances. However, instead of aggregating the features on two windows before and after, the sequential features were sampled using a sliding window at a rate of 4 Hz. This resulted in a sequence of input vectors for each utterance, containing the corresponding feature values at each instance, i.e., $x=[x_1, x_2, ..., x_\tau]$, where $\tau$ is the length of the sequence between 2 seconds before and after the end of an utterance, and $x_i \in \mathbb{R}^N$, with $N$ the total number of features at each instance.

We used two recurrent models to process the sequential data: Long Short Term Memory (LSTM) and Gated Recurrent Unit (GRU) networks. The LSTM and GRU models had one recurrent layer. As a recurrent activation function, we used the sigmoid function, and as activation for the hidden state, the hyperbolic tangent activation function. At the final time step of the recurrent models, the hidden state was fed to a dense layer with a softmax activation, resulting in a probability distribution over the possible classes. The models were trained with a categorical cross-entropy loss.  %The output vector had the same format as in the MLPs, where each value corresponded to one of the three classes used in the annotation. %Similar to the MLPs, the true distribution was the one-hot encoding vector. 
To prevent overfitting, we applied a dropout rate of 0.1 in the input and recurrent connections within the neural networks. The dimensionality of the internal state and the batch size were tuned (from values 8, 16, and 32), and the models were trained with the Adam optimizer.

%To prevent overfitting and improve the performance of the models to generalize, we applied a dropout rate of 0.1 in the input and recurrent connections within the networks. The dimensionality of the internal state and the batch size were tuned (from values 8, 16, and 32), and the models were trained with the Adam optimizer.

\subsection{Baselines}

\label{sec:baselines}
\subsubsection{Feature-Based Heuristic}
As a baseline method, we used a set of simple rule-based heuristics which use two threshold values each to predict the three classes. We chose to use a set of heuristics to provide a more general insight into the promise of feature-based heuristics. To choose the features for the heuristics, we performed a forward greedy feature selection approach on the aggregated features with an SVM with RBF kernel~\cite{ferri_comparative_1994}.
We chose to consider the three top features\footnote{For all sessions combined, the top three features were: (1) maximum speech energy, (2) speaker's head direction, and (3) standard deviation of speech energy} for the feature-based heuristics given the performance improve during the feature selection process. %The method was separately applied to the three most important features according to a forward greedy feature selection approach on the aggregated features \cite{ferri_comparative_1994}. As an estimator for the feature selection, we used SVMs with RBF kernel. %The procedure stopped after obtaining the top three features As an estimator, we used SVMs with RBF kernel. The procedure stopped after obtaining the top three features
%{\color{red}explain here how} 
%because they had the most substantial performance increase.

We computed the thresholds for each feature-based heuristic separately, maximizing the F1 score on the training set. %For each of the three features, the utterances were assigned to one of the three classes (``not appropriate", ``appropriate", and ``needed") depending on the feature value in relation to two thresholds. The threshold values for each feature were selected based on the combination that resulted in the highest F1-score on the training set. 
The final score for the feature-based baseline, as shown in the comparison tables, was calculated as the mean and standard deviation over the performance of the three feature-based heuristics on the test set and holdout set.

% Depending on a data value for a feature $f_j$ in relation to two thresholds $T_{1j}$ and $T_{2j}$, utterances were assigned to the corresponding class. Thus, for feature $j$ of utterance $i$, denoted as $f_{ij}$, the class $C_i$ is determined as,
% \begin{equation}
%     C_i =
%    \begin{cases} 
%       class_1 & \text{if } f_{ij} < T_{1j} \\
%       class_2 & \text{if } T_{1j} \leq f_{ij} \leq T_{2j} \\
%       class_3 & \text{if } T_{2j} < f_{ij}
%    \end{cases}
% \end{equation}
% where for threshold values $T_{1j}$ and $T_{2j}$ applies that \mbox{$T_{1j} < T_{2j}$}, and $class_1$, $class_2$, and $class_3$ correspond to the three classes used during annotation. Note that we tested all the combinations for classes to pair of threshold values. The threshold values for each feature were selected based on the combination that resulted in the highest overall F1-score on the training set. 
% % The feature selection method is described later in section~\ref{sec:feature-selection}. 
% The final score was determined by calculating the mean and standard deviation over the three features on the test sets.

\subsubsection{Speech-and-Pause-Based (SPB) Heuristic}
\label{sec:baseline-2}
As an additional comparison, we used a similar method to that employed by the robot during the data collection. In contrast with the three classes from the annotation process, the robot's method was used to decide only whether to change the topic. Consequently, for comparison, the method was used to classify the following classes: ``not appropriate" and the combination of the other two classes, denoted as ``appropriate/needed". 
%However, as discussed later in section~\ref{sec:discussion-conclusion}, the heuristic does not explicitly evaluate the appropriateness of the decision but rather follows a more simplified approach.

Specifically, an utterance was classified as ``appropriate/needed" if the total duration of speech in the current topic was at least 60 seconds, and was followed by a silent pause of at least 2 seconds. Otherwise, the utterance was classified as ``not appropriate". Both thresholds were experimentally determined by an expert.

%for threshold $T_{speech}$ of 60 seconds of total speech duration in the current topic and threshold $T_{pause}$ of silent pause of 2 seconds, an utterance $i$ was classified as follows:
% \begin{equation}
% C_i =
%    \begin{cases} 
%       \text{``appropriate/needed"} & \text{if } S_i \geq T_{speech} \text{ and } \\ & P_i \geq T_{pause} \\
%       \text{``not appropriate"} & \text{otherwise}
%    \end{cases}
% \end{equation}
% where $S_i$ denotes the total duration of speech on the current topic, and $P_i$ is the duration of pause after the utterance.

\section{Experiments}

\subsection{Evaluation Procedure}
Our evaluation procedure 
aimed to answer the following questions:

\begin{description}[align=left,leftmargin=0em,labelsep=0.2em,font=\textbf,itemsep=0em,parsep=0.3em]
\item[Q1. ] \textit{Can machine learning models be used to predict when the topic of the discussion should change? Can they generalize to unseen groups?} We investigated this question in the context of 
\textbf{multi-class classification}, where the models for the sequential data, the models for the non-sequential data and the feature-based heuristic made predictions over three classes (``not~appropriate", ``appropriate", and ``needed"). To better understand the effect of group size on model performance, we analyzed results based on whether the models were trained using all the features for the sessions with two participants, three participants, or all sessions together. Additionally, to investigate the generalization performance of the models, we used two different test sets: one set evaluating the performance of the same groups that are used for training data (test set) and one set with groups that were completely hold-out from training (holdout set).%The Feature-Based heuristic was used as a baseline method to determine any performance gain from the learning methods.

\item[Q2. ] \textit{Does two-step classification improve model performance compared to multi-class classification?} We investigated this question by evaluating model performance on \textbf{two-step binary classification}.
%\textbf{pairwise classification}.  %Instead of Multi-class Classification, the choice of using two-step classification was motivated by the possibility of unbalanced performance between the three classes. Thus, 
We transformed the problem of 3-class classification (as in Q1) into two binary classification problems: (1) classify ``not appropriate" vs. the combined class ``appropriate/needed"; and (2) classify ``appropriate" vs. ``needed". The former binary classification problem was the same setup used by speech-and-pause-based heuristic during the data collection process, as explained in Section~\ref{sec:baselines}. The latter binary classification problem focused on a more subtle class distinction in order to evaluate if the proposed models could predict the urgency of topic changes. 
%Indicatively, we compared the performance per class of a RF classifier in multi-class versus two-step classification, chosen for its overall good performance. Additionally, we used the Speech-and-Pause-Based heuristic as a baseline.

\item[Q3.] \textit{Which features help a robot decide when the discussion should change?} % to introduce a new topic, continue probing, or wait for the discussion to finish?}\\
To better understand the value of different features, we evaluated \textbf{classification performance using sets of features}: (1) acoustic features, (2) Kinect-derived features, (3) the 20 most important features according to feature selection, and (4) all the available features. To obtain the top 20 features, the greedy feature selection procedure explained for Feature-Based Heuristic in Section~\ref{sec:baselines} was applied for 20 features. We chose the top 20 features since there were no notable performance improvements after that. %The models were trained separately using sessions with two or three participants and all sessions together. %The results showed similar patterns in all cases; thus, we report only those of the models trained with all sessions. 
%
%The features used in feature set 3 were selected based on a forward greedy feature selection approach on the aggregated features \cite{ferri_comparative_1994}. As an estimator, we used SVMs with RBF kernel. The procedure stopped when the number of features reached 20 since we observed no particular performance improvement. Only the data from the training set was used for this procedure.
\end{description}

\subsection{Data Splitting and Balancing}
\label{sec:splitting-balancing}
The data was partitioned into training, test, and holdout sets. First, a complete session was randomly selected as a holdout set and was balanced by randomly undersampling the majority class. The test set consisted of randomly sampling utterances from the rest of the sessions (each class was determined by the 20\% of the size of the minority class). For the training set, the remaining utterances were balanced by selective oversampling of the minority class and randomly undersampling the majority class such that there were an equal number of examples per class. % to have the same size. 
More specifically, the implementation of selective oversampling used the most important feature identified by the forward greedy feature selection (as explained for the Feature-Based Heuristic in Section~\ref{sec:baselines}): only samples whose selected feature values were between the 25th and 75th percentile were considered during oversampling. Selective oversampling was favored compared to random oversampling to minimize the influence of outliers during training. It is important to note that the test set contains the unseen utterances from groups also used during training; on the other hand, the holdout session quantifies the generalization capabilities to unseen groups.

\subsection{Feature Standardization and Normalization}
In order to avoid individual differences among participants, all the features were first standardized using Z-score normalization for each participant. Then, Min-Max normalization was applied so that all the data was scaled in the range between -1 and 1. Normalization parameters were calculated from the training set only to prevent any data leaking.

\subsection{Training and Hyper-parameter Optimization}
\label{sec:training-and-parameter-optimization}
The hyper-parameters of each model were tuned using grid search. This involved performing 5-fold cross-validation, resulting in five models for each set of parameters. The combination of parameters that achieved the highest mean F1-score on the validation folds was selected, and the corresponding models were later evaluated on the test and hold-out sets by taking the mean and standard deviation of their F1-scores. For MLPs, LSTMs, and GRUs, the validation fold was also used for early stopping during the training.

\section{RESULTS}

\begin{table}[t!]
\caption[F1-scores for three-class classification.]{Mean F1-score and standard deviation on the test set, and, in parentheses, on holdout set, for the \textbf{Multi-class Classification}, separately for the different sessions.}
\label{tab:single-classifier}
\centering
\footnotesize
\begin{tabular}{p{0.5cm}p{0.7cm}P{1.7cm}P{1.7cm}P{1.7cm}}
\cline{1-5}
\multicolumn{2}{c}{} & 2 participants & 3 participants & all sessions \\ \hline

\vspace{3pt} \parbox[t]{2mm}{\multirow{5}{*}{\rotatebox[origin=c]{90}{\specialcell{non-sequential}}}}
& \vspace{0pt} DT   & 0.52±0.04 (0.50±0.06) & 0.50±0.05 (0.37±0.07) & 0.50±0.02 (0.46±0.07) \\ \cline{2-5}
& \vspace{0pt} RF   & \textbf{0.59±0.01} (0.54±0.02) & 0.49±0.01 (0.45±0.07) & 0.52±0.01 (0.46±0.03) \\ \cline{2-5}
& \vspace{0pt} SVM  & 0.49±0.01 (0.42±0.01) & 0.49±0.03 (0.40±0.05) & 0.44±0.01 (0.43±0.02) \\ \cline{2-5}
& \vspace{0pt} MLP  & 0.54±0.01 (0.38±0.07) & \textbf{0.53±0.06} (0.38±0.08) & 0.51±0.04 (0.47±0.02) \\ \hline%\specialrule{.15em}{0em}{0em}

\parbox[t]{2mm}{\multirow{2}{*}{\rotatebox[origin=c]{90}{ \specialcell{sequential \hspace{0.1pt} }}}} 
& \vspace{0pt} LSTM & 0.58±0.02 (0.41±0.06) & 0.51±0.04 (0.42±0.05) & 0.50±0.04 (0.47±0.04) \\ \cline{2-5}
& \vspace{0pt} GRU  & \textbf{0.59±0.01}  (0.45±0.02) & 0.47±0.04 (0.44±0.06) & \textbf{0.54±0.00} (0.42±0.02) \\ \hline
\end{tabular}

% baseline
\vspace{0.2cm}
\begin{tabular}{p{0.5cm}p{0.7cm}P{1.7cm}P{1.7cm}P{1.7cm}}
\hline
\vspace{-7pt} \specialcell{\hspace{7pt} Feature-Based \\ \hspace{-5pt} \textbf{Heuristic}} & & 0.48±0.11 (0.40±0.04) & 0.40±0.08 (0.41±0.07) & 0.50±0.06 (0.43±0.04) \\ \hline
\end{tabular}
\end{table}

%We provide an analysis of the performance scores of the sequential and non-sequential models in predicting the three different classes (multi-class classification). In addition, we compare the performance per class in this multi-class classification and pairwise classification achieved by first collapsing the appropriate and needed class and subsequently considering these two classes in a separate classification task. Further, we compare the results to the two baseline methods. Lastly, we analyze the importance of the different features in various models using pairwise classification.

\subsection{Multi-class Classification}
Addressing Q1, Table~\ref{tab:single-classifier} shows the mean F1-score and standard deviation on the test set and, in parenthesis, on the holdout set using multi-class classification. 
With few exceptions on test and holdout sets, the sequential and non-sequential ML models tended to outperform the Feature-Based Heuristic. Table~\ref{tab:single-classifier} does not include the Speech-and-Pause-Based Heuristic because this heuristic is applicable only to binary classification.

%In regards to comparing sequential and non-sequential approaches, 
%Although the sequential models had slightly higher scores than the non-sequential approaches in Table~\ref{tab:single-classifier}, there were no notable differences.
%In the test set, RFs and GRUs performed the best for the two-participant sessions. GRUs obtained the highest accuracies in the test set of all sessions.  However, the highest accuracies were obtained from MLPs for the test set of three-participant sessions. Concerning the holdout set, all models showed a slight decline in score and increase in standard deviation compared to the test set, yet the accuracies were still comparable. 
In general, the sequential and non-sequential models had similar F-1 performance in Table 1. For the test set, the results were mixed, with the RF and GRU having slightly better performance than other models for 2 participants, the MLP being slightly better for 3 participants, and the GRU being slightly better for all sessions. For the holdout set, the RF slightly outperformed other models.

\begin{table}[t!]
\caption[F1-scores for pairwise classification.]{Mean F1-score on the test set of all sessions for \textbf{Two-step binary classification}, SPB: speech-and-pause based heuristic.}
\label{tab:pairwise compare}
\centering
\footnotesize
\begin{tabular} { c c c c c c c }
\multicolumn{7}{c}{``not appropriate" vs.  ``appropriate/needed"}\\\hline\hline
DT & RF & SVM & MLP & LSTM & GRU & SPB heuristic \\
0.70 & 0.74 & 0.68 & 0.70 & 0.74 & \textbf{0.75} & 0.54 \\
\hline
\end{tabular}\vspace{0.5em}
\begin{tabular} { c c c c c c}
\multicolumn{6}{c}{``appropriate" vs.  ``needed"}\\\hline\hline
DT & RF & SVM & MLP & LSTM & GRU \\
0.58 & \textbf{0.61} & 0.59 & 0.57 & 0.57 & 0.58 \\
\hline
\end{tabular}
\vspace{-1em}
\end{table}

\subsection{Two-step Binary Classification}

To allow for comparison to multi-class classification in Q2, the mean F1-score of two-step binary classification for the test set of all sessions is provided in Table~\ref{tab:pairwise compare}. For the ``not appropriate" vs.  ``appropriate/needed" classification, learning-based methods %significantly 
outperformed the Speech-and-Pause-Based Heuristic. The best models for binary classification on ``not appropriate" vs. ``appropriate/needed" were the GRU for sequential models and RF for non-sequential models, with the GRU having a slightly higher F-1 score (0.75 vs. 0.74).
%where the best performance was obtained by the GRUs. 

For the ``appropriate" vs.  ``needed" classification in Table~\ref{tab:pairwise compare}, there was an %significant 
accuracy drop for the models using sequential and non-sequential data. In this classification, RF achieved the highest accuracy (0.61 average F-1 score), followed by SVMs (0.59). No heuristic approach was applicable to this classification setting, so they are omitted in Table~\ref{tab:pairwise compare}.

\subsection{Multi-class vs Two-step Binary Classification}

\begin{table}[t!]

\caption[Mean F1-score for each class category using \textbf{Two-step binary classification}.]{Mean F1-score of \textbf{each class category} on the test set of all sessions using \textbf{Two-step binary classification} - reported for RF. Results for ``not appropriate vs. appropriate/needed" on the left and for ``appropriate vs. needed" on the right.}
%\caption[Mean F1-score per class using two classifiers.]{Mean F1-score per class on the test set using \mbox{\textbf{two classifiers} (RF)} (all~sessions).
%\\Classifier (LEFT) for classes ``not appropriate" vs. ``appropriate/needed" and classifier (RIGHT) for classes ``appropriate" vs. ``needed".
%}
\label{tab:pairwise-per-class}
\begin{tabular}{lP{1.3cm}}
\hline
Class label        & F1-score  \\ \hline
not appropriate    & 0.73±0.01 \\
appropriate/needed & 0.74±0.02 \\ \hline
F1-score (macro)   & 0.74±0.02 \\ \hline
\end{tabular}
\hfill
\begin{tabular}{lP{1.3cm}}
\hline
Class label      & F1-score  \\ \hline
appropriate      & 0.65±0.03 \\
needed           & 0.58±0.05 \\ \hline
F1-score (macro) & 0.61±0.03 \\ \hline
\end{tabular}
\caption[Mean F1-score for each class category using \textbf{speech-and-pause-based heuristic}.]{Mean F1-score of \textbf{each class category} (``not appropriate vs. appropriate/needed") on the test set of all sessions using \textbf{speech-and-pause-based heuristic}.}
%\caption[Mean F1-score using speech-and-pause-based heuristic.]{Mean F1-score per class on the test set using \textbf{speech-and-pause-based heuristic} (all sessions).}
\label{tab:robot-rule-based-method}
\centering
\begin{tabular}{lP{2.5cm}}
\hline
Class label         & F1-score  \\ \hline
not appropriate     &  0.72     \\
appropriate/needed  &  0.37     \\ \hline
F1-score (macro)    &  0.54     \\ \hline
\end{tabular}
\caption[Mean F1-score for each class category using a single \textbf{Multi-Class Classification}.]{Mean F1-score of \textbf{each class category} (``not appropriate vs. appropriate vs. needed") on the test set of all sessions using \textbf{Multi-Class Classification} - reported for RF.}
\label{tab:single-classifier-per-class}
\centering
\begin{tabular}{lP{2.5cm}}
\hline
Class label         & F1-score  \\ \hline
not appropriate     & 0.64±0.02 \\
appropriate         & 0.56±0.03 \\
needed              & 0.36±0.04 \\ \hline
\end{tabular}
\vspace{-1em}
\end{table}

\begin{table*}[ht!]
\begin{minipage}{0.5\textwidth}
\caption[F1-scores for ``not appr." vs. ``appr./needed" classification.]{Mean F1-score and standard deviation on the all sessions test set, and in parentheses on holdout set, for \mbox{\textbf{``not appropriate" vs. ``appropriate/needed"}} classes, using \textbf{Two-step Binary Classification}.}
\label{tab:two-classifiers-sessions-all-1}
\centering
\begin{tabular}{p{0.7cm}P{1.4cm}P{1.4cm}P{1.4cm}P{1.4cm}}
\hline
\textbf{} & \textbf{Acoustic} & \textbf{Kinect} & \textbf{Top-20} & \textbf{All} \\ \hline
\vspace{0pt} DT   & 0.72±0.02 (0.78±0.03) & 0.54±0.03 (0.54±0.04) & 0.72±0.01 (0.75±0.02) & 0.70±0.03 (0.77±0.02) \\ \hline
\vspace{0pt} RF   & 0.74±0.02 (0.76±0.02) & 0.58±0.02 (0.56±0.06) & 0.74±0.01 (0.76±0.02) & 0.74±0.02 (0.79±0.01) \\ \hline
\vspace{0pt} SVM  & 0.73±0.01 (0.78±0.01) & 0.54±0.01 (0.54±0.02) & 0.72±0.01 (0.75±0.00) & 0.68±0.01 (0.72±0.01) \\ \hline
\vspace{0pt} MLP  & 0.73±0.01 (0.78±0.01) & 0.56±0.02 (0.56±0.03) & 0.72±0.02 (0.76±0.01) & 0.70±0.03 (0.74±0.03) \\ \hline
\vspace{0pt} LSTM  & 0.74±0.01 (0.78±0.02) & 0.58±0.02 (0.51±0.05) & 0.75±0.02 (0.75±0.01) & 0.74±0.03 (0.74±0.02) \\ \hline
\vspace{0pt} GRU  & \textbf{0.76±0.01} (0.77±0.02) & 0.61±0.02 (0.49±0.04) & \textbf{0.76±0.01} (0.76±0.01) & 0.75±0.01 (0.74±0.03) \\ \hline
\end{tabular}
\end{minipage}
\hfill
\begin{minipage}{0.5\textwidth}
\caption[F1-scores for ``appr." vs. ``needed" classification.]{Mean F1-score and standard deviation on the all sessions test set, and in parentheses on holdout set, for \mbox{\textbf{``appropriate" vs. ``needed"}} classes, using \textbf{Two-step Binary Classification}.}
\label{tab:two-classifiers-sessions-all-2}
\begin{tabular}{p{0.7cm}P{1.4cm}P{1.4cm}P{1.4cm}P{1.4cm}}
\hline
\textbf{} & \textbf{Acoustic} & \textbf{Kinect} & \textbf{Top-20} & \textbf{All} \\ \hline

\vspace{0pt} DT  & 0.59±0.03 (0.54±0.04) & 0.54±0.03 (0.51±0.03) & 0.57±0.04 (0.47±0.06) & 0.58±0.06 (0.53±0.06) \\ \hline
\vspace{0pt} RF  & 0.60±0.01 (0.57±0.06) & 0.54±0.03 (0.46±0.05) & 0.57±0.04 (0.53±0.09) & \textbf{0.61±0.03} (0.50±0.05) \\ \hline
\vspace{0pt} SVM  & 0.59±0.02 (0.58±0.03) & 0.56±0.02 (0.47±0.02) & 0.51±0.03 (0.55±0.02) & 0.59±0.02 (0.50±0.03) \\ \hline
\vspace{0pt} MLP  & 0.58±0.03 (0.57±0.05) & 0.56±0.04 (0.47±0.03) & 0.54±0.05 (0.50±0.09) & 0.57±0.02 (0.45±0.05) \\ \hline
\vspace{0pt} LSTM  & \textbf{0.61±0.02} (0.57±0.06) & 0.53±0.02 (0.49±0.05) & 0.55±0.03 (0.44±0.03) & 0.57±0.03 (0.50±0.04) \\ \hline
\vspace{0pt} GRU  & 0.59±0.03 (0.54±0.04) & 0.50±0.01 (0.42±0.08) & 0.51±0.06 (0.44±0.04) & 0.58±0.02 (0.45±0.05) \\ \hline
\end{tabular}
\end{minipage}
\vspace{-1em}
\end{table*}

In the previous sections, we reported the aggregate performance for Multi-class and Two-step Binary classifiers. To answer Q2, we compare these results considering each class category. We use the SPB Heuristic as a baseline. The results for multi-class classification (using RF), two-step binary classification (using RF), and the heuristic are provided in Tables~\ref{tab:single-classifier-per-class}, ~\ref{tab:pairwise-per-class} and ~\ref{tab:robot-rule-based-method}, respectively.

Compared to the SPB Heuristic, two-step binary classification by the RF model performs %significantly 
better for each class. Even though the heuristic had a relatively high score in the ``not appropriate" class, it performed poorly on determining when a topic change is ``appropriate/needed". For this category, the two-step binary approach using RF %pairwise approach 
obtained $\sim$37\% higher accuracy than the heuristic approach (Two-step Binary RF Model: \%0.74±0.02, SPB heuristic:\%0.37).

Lastly, when we compare one-step (multi-class classification) and two-step approaches (two-step binary classification), both models showed high performances in classifying the ``not appropriate" class. Although both methods used RF as a classifier, there was a notable performance drop for the ``appropriate'' vs ``needed" classes when the problem was formulated with a single-step approach.

%Tables~\ref{tab:single-classifier-per-class} and~\ref{tab:pairwise-per-class} present the scores for each class using only one and two classifiers, respectively. In the case of multi-class classification (Table~\ref{tab:single-classifier-per-class}), the models performed better in classifying the ``not appropriate" class, with the other two classes being noticeably lower. The pairwise classification resulted in comparable results in each pair of classes. However, the individual performance for the classes ``appropriate" and ``needed" remained low (see Table~\ref{tab:pairwise-per-class} (RIGHT)).

%As an additional comparison, Table~\ref{tab:robot-rule-based-method} presents the F1-score per class using the speech-and-pause-based heuristic (section~\ref{sec:baseline-2}). The results are reported only for the RF classifier and the speech-and-pause-based heuristic because...{\color{red} we should explain here, old explanation that I couldn't understand: Indicatively, we compared the performance per class of a RF classifier in multi-class versus two-step classification, chosen for its overall good performance. Additionally, we used the Speech-and-Pause-Based heuristic as a baseline.}

%Regarding the performance of the heuristic, even though the method had a relatively high score in the ``not appropriate" class, it performed poorly on determining when a topic change is ``appropriate/needed".

\subsection{Classification Performance Using Sets of Features}

To investigate Q3, we provide the accuracy results of each model using varying sets of input features instead of using the whole feature set. The results were obtained using the Two-step Binary classification given the promise of this approach explored in Q2.
Results are presented as the mean F1-score and standard deviation on the test set of all sessions and the holdout set (in parenthesis) in Table~\ref{tab:two-classifiers-sessions-all-1} and  ~\ref{tab:two-classifiers-sessions-all-2}.

%The results showed similar patterns in all cases; thus, we report only those of the models trained with all sessions. The models were trained separately using sessions with two or three participants and all sessions together.  The features used in feature set 3 were selected based on a forward greedy feature selection approach on the aggregated features \cite{ferri_comparative_1994}. As an estimator, we used SVMs with RBF kernel. The procedure stopped when the number of features reached 20 since we observed no particular performance improvement. Only the data from the training set was used for this procedure.

%The results are presented as the mean F1-score and standard deviation on the test set and corresponding values in parenthesis on the holdout set. Table~\ref{tab:two-classifiers-sessions-all-1} corresponds to the classes ``not appropriate" vs. ``appropriate/needed", and Table~\ref{tab:two-classifiers-sessions-all-2} to ``appropriate" vs. ``needed".

% features

In the test set (Table~\ref{tab:two-classifiers-sessions-all-1}), the performance of the acoustic and Top-20 features was slightly higher than using all the features, with the GRU performing the best (0.76~±~0.1 for acoustic and Top-20 features vs. 0.75~±~0.01 for all features). For the holdout set, the best results were obtained with all features and the acoustic features. In particular, the RF model had a slightly higher performance with all the features (0.79~±~0.01) than several other models with acoustic features (which reached 0.78 average F-1 scores). Thus, the main takeaway from these results is that using only Kinect body features to distinguish between ``not appropriate" and ``appropriate/needed" is worse than incorporating other features into this prediction problem. Surprisingly, acoustic features often led to good performance in this classification task.
%Concerning the type of features, the performance of models using only acoustic and Top-20 features was comparable or slightly higher than the performance for all features. Using the Kinect-derived features had the lowest performance in both test and holdout set. %, and slightly increased standard deviation. %The performance using the top-20 features was comparable to all features and showed improvements in certain cases, but had overall slightly lower performance than with the acoustic features.
%Concerning the type of features, the performance of models using only acoustic features was comparable (or even higher) to all features. Using the Kinect-derived features had the lowest performance in both test and holdout set, and slightly increased standard deviation. The performance using the top-20 features was comparable to all features and showed improvements in certain cases, but had overall slightly lower performance than with the acoustic features.
% models
%Regarding the machine learning models, although LSTMs and GRUs yielded slightly better performances, there were not any notable performance variations between models. 

All the models showed a decrease while classifying ``appropriate" vs. ``needed" (Table~\ref{tab:two-classifiers-sessions-all-2}) compared to classifying ``not appropriate" vs. ``appropriate/needed" (Table~\ref{tab:two-classifiers-sessions-all-1}), with no apparent correlations using different sets of features. Finally, holdout set accuracies (in both tables) showed similar trends with all session test set performances.

%Regarding the machine learning models, there was not any significant/consistent performance increase using sequential models (LSTMs and GRUs) compared to the rest. Furthermore, there was no clear pattern as to which model performed best.

% classification
%In terms of the classification problem, there was an overall noticeable decrease in the performance of all models in classifying ``appropriate" vs. ``needed", with no apparent difference between the models trained with different features.
\section{DISCUSSION}
%\subsection{Effectiveness of Machine Learning Models}
In this study, we investigated the effectiveness of machine learning models in topic change prediction using non-verbal features. Given the complexity of the task, instead of following heuristic-based decisions, we suggest the need for learning-based methods for robots to be capable of topic moderation. Accordingly, we evaluate various ML models using sequential and non-sequential inputs, and we provide further analysis using one-step multi-class and two-step binary classification techniques. 
Our findings suggest the applicability of using ML approaches for topic change in robot-facilitated discussions. They also show that using acoustic data or the most informative features can provide comparable results with the whole future set.
%Our findings show the advantages of ML models over heuristic-based approaches. Additionally, we provide an extensive analysis of feature selection, which shows the importance of selecting the most informative features in the decision process as well as the acoustic data. 
%We initially evaluated the two types of models in multi-class classification with learning approaches to achieve higher performance compared to the feature-based heuristic, suggesting the need for learning methods for robots for the task of topic moderation. However, considering the performance of the heuristic, which used only three features, there is a strong indication that fewer features were needed for the task. 
This can guide future HRI research to simplify features used without compromising the prediction performances.
%performance of the models in the task. 

While exploring Q1 and Q2, we compared a sensible heuristic against varying ML models (Table~\ref{tab:pairwise-per-class}, ~\ref{tab:robot-rule-based-method}, and ~\ref{tab:single-classifier-per-class}). Even though the heuristic method had relatively high accuracy in the ``not appropriate" class, it performed poorly on determining when a topic change is ``appropriate/needed." This highlights the lack of flexibility and effectiveness of rule-based methods compared to the learning models and further highlights the need for learning methods for this task. Additionally, relatively similar performances of ML models on the test and hold-out data show these models' robustness and generalization capabilities (Table~\ref{tab:single-classifier}).

Another finding demonstrates the robustness of ML models on unseen data obtained from the analysis for Q3 investigating varying sets of features (Table~\ref{tab:two-classifiers-sessions-all-1} and ~\ref{tab:two-classifiers-sessions-all-2}). Regarding the type of features, acoustic and Top-20 features were identified as promising choices. Kinect-derived features showed low overall performance and generalization issues. Their performance could be attributed to the higher dimensionality of the features compared to acoustic features or person-specificity. An additional reason for the low impact of Kinect features could be the discussion topics during the brainstorming sessions, which, in contrast with previous work on gestures and topic structure \cite{quek_gesture_2000, cassell_non-verbal_2001}, did not mainly involve spatial information, that could otherwise encourage using hand gestures. In addition, cultural differences are known to affect the use of gestures \cite{kendon1981geography}, which might have influenced the results for the Kinect features.

Considering the type of classification techniques explored for Q2, two-step binary classification reported higher accuracy than multi-class classification. This could be due to the chance level increase when the process was simplified to binary classification instead of making a prediction among three classes. Additionally, in binary classification between ``not appropriate vs. appropriate/needed"  obtained higher accuracy than ``appropriate vs. needed". This suggests that the decision is easier for the robot when it is not expected to change the discussion topic (higher accuracy for ``not appropriate vs. appropriate/needed" prediction). However, when adding the possibility to change the topic, the decision is harder (lower accuracy for ``appropriate vs. needed").

%has an opportunity to change the topic (class ``appropriate") and where it was obliged to do so (class ``needed") is a more difficult challenge. However, classifying between the time a topic change was not appropriate at all and when a new topic could be introduced led to promising results, demonstrating the feasibility of topic moderation via machine learning methods.

Regarding the type of ML models, there was no clear advantage in using sequential approaches over non-sequential approaches. These findings indicate that the aggregated features could provide enough information and, while combined with simpler models, achieve comparable results without the complexity of the sequential method.  This finding is especially interesting for HRI contexts as simpler models also have lower data requirements, beneficial given the cost and complexity of collecting human-robot interaction data. %Moreover, utilizing simpler models with fewer computational requirements can potentially increase the responsiveness of robots in deciding on topic changes in group discussions. %Sequential models 
Nonetheless, more data could benefit the models, especially for using sequential approaches, given the high dimensionality of the input; thus, the amount of data available could have affected the performance of our experiments.
%, and an additional requirement to be considered for applications in robot moderated discussions.

%Due to the robot's role as the moderator of the discussions during data collection, the ground truth was not available. Despite the use of the ``appropriate" class to account for an opportunity for a topic shift and mitigate this issue, it remains a limitation of the data that was used.
One of the main challenges of topic change in robot-facilitated discussion is benchmarking. Given the complexity of the task, there are no learning-based baselines to build upon or publicly available datasets. This motivated us to gather our own dataset, yet it has a limitation of the finite quantity of participants and interaction sessions we had at our disposal, combined with its imbalanced nature. Therefore, the aim of future research could be the collection of a more extensive and diverse dataset.
Lastly, future research could investigate how the proposed methods can be applied to larger groups, which could reveal additional opportunities and contribute to a wider social context and practical applications of robot-moderated discussions.

\section{CONCLUSION}
In this paper, we proposed a novel technical problem: when should a robot change the topic of a robot-moderated group discussion? Further, we provided insight into using machine learning approaches to solve this problem. %for robots to determine when to change the topic of group discussions.
Our results demonstrate the complexity of the task. Heuristic-based approaches under-performed learning-based methods, showing the value of machine learning in this problem domain.
%In conclusion, the current work provides insights into the models and features for effective robot-mediated topic moderation. Additional investigation of body language and its combination with acoustic features is of particular interest. 
%Also, %we showed that the feature selection process presents opportunities to improve the topic change classification. O
Our findings demonstrate the importance of selecting the most informative features when predicting topic changes and, surprisingly, suggested that acoustic features are particularly useful. Overall, our work provided a new dataset for  %makes a critical step %Finally, the limitations of the lack of ground truth and baseline methods highlight the complexity of the task and the need 
automated topic change decisions in robot-moderated group discussions and an initial exploration of models that could be used to address this technical challenge. %, and presents avenues for future research by providing detailed insights as well as the dataset.

\bibliographystyle{IEEEtran}
\bibliography{references_shortened}

\begin{thebibliography}{10}
\providecommand{\url}[1]{#1}
\csname url@samestyle\endcsname
\providecommand{\newblock}{\relax}
\providecommand{\bibinfo}[2]{#2}
\providecommand{\BIBentrySTDinterwordspacing}{\spaceskip=0pt\relax}
\providecommand{\BIBentryALTinterwordstretchfactor}{4}
\providecommand{\BIBentryALTinterwordspacing}{\spaceskip=\fontdimen2\font plus
\BIBentryALTinterwordstretchfactor\fontdimen3\font minus \fontdimen4\font\relax}
\providecommand{\BIBforeignlanguage}[2]{{%
\expandafter\ifx\csname l@#1\endcsname\relax
\typeout{** WARNING: IEEEtran.bst: No hyphenation pattern has been}%
\typeout{** loaded for the language `#1'. Using the pattern for}%
\typeout{** the default language instead.}%
\else
\language=\csname l@#1\endcsname
\fi
#2}}
\providecommand{\BIBdecl}{\relax}
\BIBdecl

\bibitem{forsyth2014group}
D.~R. Forsyth, \emph{Group dynamics}.\hskip 1em plus 0.5em minus 0.4em\relax Wadsworth Cengage Learning, 2014.

\bibitem{sebo_robots_2020}
S.~Sebo, B.~Stoll, B.~Scassellati, and M.~F. Jung, ``\BIBforeignlanguage{en}{\href{https://dl.acm.org/doi/10.1145/3415247}{Robots in {Groups} and {Teams}: {A} {Literature} {Review}}},'' \emph{\BIBforeignlanguage{en}{Proceedings of the ACM on Human-Computer Interaction}}, vol.~4, no. CSCW2, pp. 1--36, Oct. 2020.

\bibitem{ISRgillet2024}
S.~Gillet, M.~V\'{a}zquez, S.~Andrist, I.~Leite, and S.~Sebo, ``\href{https://doi.org/10.1145/3643803}{Interaction-Shaping Robotics: Robots that Influence Interactions between Other Agents},'' \emph{J. Hum.-Robot Interact.}, 2024.

\bibitem{sabanovic_paro_2013}
S.~Sabanovic, C.~C. Bennett, {Wan-Ling Chang}, and L.~Huber, ``\href{http://ieeexplore.ieee.org/document/6650427/}{PARO} robot affects diverse interaction modalities in group sensory therapy for older adults with dementia,'' in \emph{2013 {IEEE} 13th {International} {Conference} on {Rehabilitation} {Robotics} ({ICORR})}.\hskip 1em plus 0.5em minus 0.4em\relax IEEE, 2013.

\bibitem{thompson_robot_2017}
C.~Thompson, S.~Mohamed, W.-Y.~G. Louie, J.~C. He, J.~Li, and G.~Nejat, ``\href{http://ieeexplore.ieee.org/document/8250117/}{The robot {Tangy} facilitating {Trivia} games: {A} team-based user-study with long-term care residents},'' in \emph{2017 {IEEE} {International} {Symposium} on {Robotics} and {Intelligent} {Sensors} ({IRIS})}.\hskip 1em plus 0.5em minus 0.4em\relax IEEE, 2017.

\bibitem{tennent_micbot_2019}
H.~Tennent, S.~Shen, and M.~Jung, ``\BIBforeignlanguage{en}{\href{https://ieeexplore.ieee.org/document/8673013/}{Micbot: {A} {Peripheral} {Robotic} {Object} to {Shape} {Conversational} {Dynamics} and {Team} {Performance}}},'' in \emph{\BIBforeignlanguage{en}{2019 14th {ACM}/{IEEE} {International} {Conference} on {Human}-{Robot} {Interaction} ({HRI})}}.\hskip 1em plus 0.5em minus 0.4em\relax Daegu, Korea (South): IEEE, 2019.

\bibitem{Strohkorb2016}
S.~Strohkorb, E.~Fukuto, N.~Warren, C.~Taylor, B.~Berry, and B.~Scassellati, ``\href{https://ieeexplore.ieee.org/document/8673013/}{Improving human-human collaboration between children with a social robot},'' \emph{25th IEEE International Symposium on Robot and Human Interactive Communication, RO-MAN 2016}, 2016.

\bibitem{short_robot_2017}
E.~Short and M.~J. Mataric, ``\href{http://ieeexplore.ieee.org/document/8172331/}{Robot moderation of a collaborative game: {Towards} socially assistive robotics in group interactions},'' in \emph{2017 26th {IEEE} {International} {Symposium} on {Robot} and {Human} {Interactive} {Communication} ({RO}-{MAN})}.\hskip 1em plus 0.5em minus 0.4em\relax IEEE, 2017.

\bibitem{de2023co}
A.~de~Rooij, S.~van~den Broek, M.~Bouw, and J.~de~Wit, ``Co-designing with a social robot facilitator: Effects of robot mood expression on human group dynamics,'' in \emph{Proceedings of the 11th International Conference on Human-Agent Interaction}, 2023.

\bibitem{10.3389/frobt.2021.657291}
J.~Geerts, J.~de~Wit, and A.~de~Rooij, ``\href{https://www.frontiersin.org/articles/10.3389/frobt.2021.657291}{Brainstorming With a Social Robot Facilitator: Better Than Human Facilitation Due to Reduced Evaluation Apprehension?}'' \emph{Frontiers in Robotics and AI}, vol.~8, 2021.

\bibitem{mizrahi2022vrobotator}
E.~Mizrahi, N.~Danzig, and G.~Gordon, ``vrobotator: A virtual robot facilitator of small group discussions for k-12,'' \emph{Proceedings of the ACM on Human-Computer Interaction}, vol.~6, no. CSCW2, 2022.

\bibitem{kovacs_topical_2016}
G.~Kovacs, T.~Grosz, and T.~Varadi, ``\BIBforeignlanguage{en}{\href{http://ieeexplore.ieee.org/document/7804549/}{Topical unit classification using deep neural nets and probabilistic sampling}},'' in \emph{\BIBforeignlanguage{en}{7th {IEEE} {International} {Conference} on {Cognitive} {Infocommunications}}}.\hskip 1em plus 0.5em minus 0.4em\relax IEEE, 2016.

\bibitem{hunyadi_temporal_2020}
L.~Hunyadi and I.~Szekrényes, Eds., \emph{\BIBforeignlanguage{en}{\href{http://link.springer.com/10.1007/978-3-030-22895-8}{The {Temporal} {Structure} of {Multimodal} {Communication}: {Theory}, {Methods} and {Applications}}}}, ser. Intelligent {Systems} {Reference} {Library}.\hskip 1em plus 0.5em minus 0.4em\relax Springer International Publishing, 2020, vol. 164.

\bibitem{tomiyama_identifying_2018}
K.~Tomiyama, F.~Nihei, Y.~I. Nakano, and Y.~Takase, ``\BIBforeignlanguage{en}{Identifying {Discourse} {Boundaries} in {Group} {Discussions} using a {Multimodal} {Embedding} {Space}},'' in \emph{\BIBforeignlanguage{en}{{ACM} {IUI} 2018 {Workshops}, {Symbiotic} {Interaction} and {Harmonious} {Collaboration} for {Wisdom} {Computing}}}, 2018.

\bibitem{NonDyadicSchneiders22}
E.~Schneiders, E.~Cheon, J.~Kjeldskov, M.~Rehm, and M.~B. Skov, ``Non-dyadic interaction: A literature review of 15 years of human-robot interaction conference publications,'' \emph{J. Hum.-Robot Interact.}, vol.~11, no.~2, 2022.

\bibitem{Martelaro2015}
M.~F. Jung, N.~Martelaro, and P.~J. Hinds, ``\href{http://dl.acm.org/citation.cfm?doid=2701973.2702094}{Using Robots to Moderate Team Conflict: The Case of Repairing Violations},'' in \emph{Proceedings of the ACM/IEEE International Conference on Human-Robot Interaction}.\hskip 1em plus 0.5em minus 0.4em\relax Association for Computing Machinery, 2015.

\bibitem{Shen2018}
S.~Shen, P.~Slovak, and M.~F. Jung, ``\href{{https://dl.acm.org/doi/10.1145/3171221.3171248}}{"Stop. I See a Conflict Happening."},'' in \emph{Proceedings of the 2018 ACM/IEEE International Conference on Human-Robot Interaction}.\hskip 1em plus 0.5em minus 0.4em\relax ACM, 2018.

\bibitem{erel2021enhancing}
H.~Erel, D.~Trayman, C.~Levy, A.~Manor, M.~Mikulincer, and O.~Zuckerman, ``Enhancing emotional support: The effect of a robotic object on human--human support quality,'' \emph{International Journal of Social Robotics}, pp. 1--20, 2021.

\bibitem{StrohkorbSebo2018}
S.~Strohkorb~Sebo, M.~Traeger, M.~F. Jung, and B.~Scassellati, ``\href{http://dl.acm.org/citation.cfm?doid=3171221.3171275}{The Ripple Effects of Vulnerability: The Effects of a Robot's Vulnerable Behavior on Trust in Human-Robot Teams},'' \emph{Proceedings of the 2018 ACM/IEEE International Conference on Human-Robot Interaction - HRI '18}, no. February, pp. 178--186, 2018.

\bibitem{StrohkorbSebo2020StrategiesTeams}
S.~Strohkorb~Sebo, L.~L. Dong, N.~Chang, and B.~Scassellati, ``\href{https://dl.acm.org/doi/10.1145/3319502.3374808}{Strategies for the Inclusion of Human Members within Human-Robot Teams},'' in \emph{Proceedings of the 2020 ACM/IEEE International Conference on Human-Robot Interaction}.\hskip 1em plus 0.5em minus 0.4em\relax ACM, 3 2020.

\bibitem{gillet2020mediator}
S.~Gillet, W.~van~den Bos, and I.~Leite, ``{A social robot mediator to foster collaboration and inclusion among children},'' in \emph{Proceedings of Robotics: Science and Systems}, Corvalis, Oregon, USA, 2020.

\bibitem{tuncerSmileInclusion}
S.~Tuncer, S.~Gillet, and I.~Leite, ``Robot-mediated inclusive processes in groups of children: From gaze aversion to mutual smiling gaze,'' \emph{Frontiers in Robotics and AI}, vol.~9, 2022.

\bibitem{RobotLevelGilletCumbal2021}
S.~Gillet, R.~Cumbal, A.~Pereira, J.~Lopes, O.~Engwall, and I.~Leite, ``{Robot Gaze Can Mediate Participation Imbalance in Groups with Different Skill Levels},'' in \emph{Proceedings of the 2021 ACM/IEEE International Conference on Human-Robot Interaction}.\hskip 1em plus 0.5em minus 0.4em\relax ACM, 2021.

\bibitem{vazquez2015social}
M.~V{\'a}zquez, E.~J. Carter, J.~A. Vaz, J.~Forlizzi, A.~Steinfeld, and S.~E. Hudson, ``Social group interactions in a role-playing game,'' in \emph{Proceedings of the Tenth Annual ACM/IEEE International Conference on Human-Robot Interaction Extended Abstracts}, 2015.

\bibitem{vazquez2017towards}
M.~V{\'a}zquez, E.~J. Carter, B.~McDorman, J.~Forlizzi, A.~Steinfeld, and S.~E. Hudson, ``Towards robot autonomy in group conversations: Understanding the effects of body orientation and gaze,'' in \emph{Proceedings of the 2017 ACM/IEEE International Conference on Human-Robot Interaction}, 2017.

\bibitem{grassi_knowledge-grounded_2022}
L.~Grassi, C.~T. Recchiuto, and A.~Sgorbissa, ``\BIBforeignlanguage{en}{\href{https://link.springer.com/10.1007/s12369-022-00868-z}{Knowledge-{Grounded} {Dialogue} {Flow} {Management} for {Social} {Robots} and {Conversational} {Agents}}},'' \emph{\BIBforeignlanguage{en}{International Journal of Social Robotics}}, vol.~14, no.~5, 2022.

\bibitem{glas_topic_2018}
N.~Glas and C.~Pelachaud, ``\BIBforeignlanguage{en}{\href{https://linkinghub.elsevier.com/retrieve/pii/S1071581918304233}{Topic management for an engaging conversational agent}},'' \emph{\BIBforeignlanguage{en}{International Journal of Human-Computer Studies}}, vol. 120, 2018.

\bibitem{tang_target-guided_2019}
J.~Tang, T.~Zhao, C.~Xiong, X.~Liang, E.~Xing, and Z.~Hu, ``\BIBforeignlanguage{en}{\href{https://www.aclweb.org/anthology/P19-1565}{Target-{Guided} {Open}-{Domain} {Conversation}}},'' in \emph{\BIBforeignlanguage{en}{Proceedings of the 57th {Annual} {Meeting} of the {Association} for {Computational} {Linguistics}}}.\hskip 1em plus 0.5em minus 0.4em\relax Florence, Italy: Association for Computational Linguistics, 2019.

\bibitem{sevegnani_otters_2021}
K.~Sevegnani, D.~M. Howcroft, I.~Konstas, and V.~Rieser, ``\BIBforeignlanguage{en}{\href{https://aclanthology.org/2021.acl-long.194}{OTTers}: {One}-turn {Topic} {Transitions} for {Open}-{Domain} {Dialogue}},'' in \emph{\BIBforeignlanguage{en}{Proceedings of the 59th {Ann.} {Meeting} of the {Assoc.} for {Computational} {Linguistics} and the 11th {Int.} {Joint} {Conference} on {Natural} {Language} {Processing}}}.\hskip 1em plus 0.5em minus 0.4em\relax Association for Computational Linguistics, 2021.

\bibitem{rich_recognizing_2010}
C.~Rich, B.~Ponsler, A.~Holroyd, and C.~L. Sidner, ``\href{http://ieeexplore.ieee.org/document/5453163/}{Recognizing engagement in human-robot interaction},'' in \emph{5th {ACM}/{IEEE} {International} {Conference} on {Human}-{Robot} {Interaction} ({HRI})}.\hskip 1em plus 0.5em minus 0.4em\relax IEEE, 2010.

\bibitem{hutchison_direction_2005}
C.~Peters, ``\href{http://link.springer.com/10.1007/11550617{\_}19}{Direction of {Attention} {Perception} for {Conversation} {Initiation} in {Virtual} {Environments}},'' in \emph{Intelligent {Virtual} {Agents}}.\hskip 1em plus 0.5em minus 0.4em\relax Berlin, Heidelberg: Springer Berlin Heidelberg, 2005, vol. 3661, series Title: Lecture Notes in Computer Science.

\bibitem{ishii_gaze_2013}
R.~Ishii, Y.~I. Nakano, and T.~Nishida, ``\BIBforeignlanguage{en}{\href{https://dl.acm.org/doi/10.1145/2499474.2499480}{Gaze awareness in conversational agents: {Estimating} a user's conversational engagement from eye gaze}},'' \emph{\BIBforeignlanguage{en}{ACM Transactions on Interactive Intelligent Systems}}, vol.~3, no.~2, 2013.

\bibitem{eisenstein_gestural_2008}
J.~Eisenstein, R.~Barzilay, and R.~Davis, ``\BIBforeignlanguage{en}{\href{https://aclanthology.org/P08-1}{Gestural {Cohesion} for {Topic} {Segmentation}}},'' in \emph{\BIBforeignlanguage{en}{Proceedings of {ACL}-08: {HLT}}}.\hskip 1em plus 0.5em minus 0.4em\relax Columbus, Ohio: Association for Computational Linguistics, 2008.

\bibitem{nakajima_study_1993}
S.~Nakajima and J.~F. Allen, ``\BIBforeignlanguage{en}{A {Study} on {Prosody} and {Discourse} {Structure} in {Cooperative} {Dialogues}},'' \emph{\BIBforeignlanguage{en}{Phonetica}}, vol.~50, no.~3, 1993.

\bibitem{herman_phonetic_2000}
R.~Herman, ``\BIBforeignlanguage{en}{\href{https://linkinghub.elsevier.com/retrieve/pii/S009544700090127X}{Phonetic markers of global discourse structures in {English}}},'' \emph{\BIBforeignlanguage{en}{Journal of Phonetics}}, vol.~28, no.~4, 2000.

\bibitem{swerts_melodic_1994}
M.~Swerts, D.~G. Bouwhuis, and R.~Collier, ``\BIBforeignlanguage{en}{\href{http://asa.scitation.org/doi/10.1121/1.410148}{Melodic cues to the perceived ‘‘finality’’ of utterances}},'' \emph{\BIBforeignlanguage{en}{The Journal of the Acoustical Society of America}}, vol.~96, no.~4, 1994.

\bibitem{zellers_prosodic_2011}
M.~K. Zellers, ``\BIBforeignlanguage{en}{Prosodic {Detail} and {Topic} {Structure} in {Discourse}},'' Ph.D. dissertation, University of Cambridge UK, 2011.

\bibitem{quek_gestural_2002}
F.~Quek, Y.~Xiong, and D.~McNeill, ``\BIBforeignlanguage{en}{\href{https://www.isca-speech.org/archive/icslp_2002/quek02_icslp.html}{Gestural trajectory symmetries and discourse segmentation}},'' in \emph{\BIBforeignlanguage{en}{7th {International} {Conference} on {Spoken} {Language} {Processing} ({ICSLP} 2002)}}.\hskip 1em plus 0.5em minus 0.4em\relax ISCA, 2002.

\bibitem{duncan_signals_1972}
S.~Duncan, ``\BIBforeignlanguage{en}{\href{http://doi.apa.org/getdoi.cfm?doi=10.1037/h0033031}{Some signals and rules for taking speaking turns in conversations.}}'' \emph{\BIBforeignlanguage{en}{Journal of Personality and Social Psychology}}, vol.~23, no.~2, 1972.

\bibitem{zellers_prosody_2016}
M.~Zellers, D.~House, and S.~Alexanderson, ``\BIBforeignlanguage{en}{\href{https://www.isca-archive.org/speechprosody_2016/zellers16_speechprosody.html}{Prosody and hand gesture at turn boundaries in {Swedish}}},'' in \emph{\BIBforeignlanguage{en}{Speech {Prosody} 2016}}.\hskip 1em plus 0.5em minus 0.4em\relax ISCA, 2016.

\bibitem{auer_previews_1992}
J.~Streeck and U.~Hartge, ``\BIBforeignlanguage{en}{\href{https://benjamins.com/catalog/pbns.22.10str}{Previews: {Gestures} at the {Transition} {Place}}},'' in \emph{\BIBforeignlanguage{en}{Pragmatics \& {Beyond} {New} {Series}}}, P.~Auer and A.~Di~Luzio, Eds.\hskip 1em plus 0.5em minus 0.4em\relax John Benjamins Publishing Company, 1992, vol.~22.

\bibitem{cassell_non-verbal_2001}
J.~Cassell, Y.~I. Nakano, T.~W. Bickmore, C.~L. Sidner, and C.~Rich, ``\BIBforeignlanguage{en}{\href{http://portal.acm.org/citation.cfm?doid=1073012.1073028}{Non-verbal cues for discourse structure}},'' in \emph{\BIBforeignlanguage{en}{Proceedings of the 39th {Annual} {Meeting} on {Association} for {Computational} {Linguistics} - {ACL} '01}}.\hskip 1em plus 0.5em minus 0.4em\relax Association for Computational Linguistics, 2001.

\bibitem{goodwin_concurrent_1987}
C.~Goodwin and M.~H. Goodwin, ``Concurrent {Operations} on {Talk}: {Notes} on the {Interactive} {Organization} of {Assesments},'' \emph{IPrA papers in pragmatics}, vol.~1, no.~1, pp. 1--54, Jan. 1987.

\bibitem{sidnell_gaze_2012}
F.~Rossano, ``\BIBforeignlanguage{en}{\href{{https://onlinelibrary.wiley.com/doi/10.1002/9781118325001.ch15}}{Gaze in {Conversation}}},'' in \emph{\BIBforeignlanguage{en}{The {Handbook} of {Conversation} {Analysis}}}, 1st~ed., J.~Sidnell and T.~Stivers, Eds.\hskip 1em plus 0.5em minus 0.4em\relax Wiley, 2012.

\bibitem{quek_gesture_2000}
F.~Quek, D.~McNeill, R.~Bryll, C.~Kirbas, H.~Arslan, K.~McCullough, N.~Furuyama, and R.~Ansari, ``\BIBforeignlanguage{en}{\href{http://ieeexplore.ieee.org/document/854800/}{Gesture, speech, and gaze cues for discourse segmentation}},'' in \emph{\BIBforeignlanguage{en}{Proceedings {IEEE} {Conference} on {Computer} {Vision} and {Pattern} {Recognition}. {CVPR} 2000 ({Cat}. {No}.{PR00662})}}, vol.~2.\hskip 1em plus 0.5em minus 0.4em\relax IEEE Comput. Soc, 2000.

\bibitem{10.1145/3610978.3641090}
S.~Thompson, A.~Narcomey, A.~Lew, and M.~V\'{a}zquez, ``\href{https://doi.org/10.1145/3610978.3641090}{Shutter: A Low-Cost and Flexible Social Robot Platform for In-the-Wild Deployments},'' in \emph{Companion of the 2024 ACM/IEEE International Conference on Human-Robot Interaction}, ser. HRI '24.\hskip 1em plus 0.5em minus 0.4em\relax Association for Computing Machinery, 2024.

\bibitem{skantze_turn-taking_2021}
G.~Skantze, ``\BIBforeignlanguage{en}{\href{https://linkinghub.elsevier.com/retrieve/pii/S088523082030111X}{Turn-taking in {Conversational} {Systems} and {Human}-{Robot} {Interaction}: {A} {Review}}},'' \emph{\BIBforeignlanguage{en}{Computer Speech \& Language}}, vol.~67, 2021.

\bibitem{gravano_turn-taking_2011}
A.~Gravano and J.~Hirschberg, ``\BIBforeignlanguage{en}{\href{{https://linkinghub.elsevier.com/retrieve/pii/S0885230810000690}}{Turn-taking cues in task-oriented dialogue}},'' \emph{\BIBforeignlanguage{en}{Computer Speech \& Language}}, vol.~25, no.~3, 2011.

\bibitem{johansson_opportunities_2015}
M.~Johansson and G.~Skantze, ``\BIBforeignlanguage{en}{\href{{http://aclweb.org/anthology/W15-4642}}{Opportunities and {Obligations} to {Take} {Turns} in {Collaborative} {Multi}-{Party} {Human}-{Robot} {Interaction}}},'' in \emph{\BIBforeignlanguage{en}{Proceedings of the 16th {Annual} {Meeting} of the {Special} {Interest} {Group} on {Discourse} and {Dialogue}}}.\hskip 1em plus 0.5em minus 0.4em\relax Prague, Czech Republic: Association for Computational Linguistics, 2015.

\bibitem{nair_rectified_2010}
V.~Nair and G.~E. Hinton, ``Rectified {Linear} {Units} {Improve} {Restricted} {Boltzmann} {Machines},'' in \emph{Proceedings of the 27th {International} {Conference} on {International} {Conference} on {Machine} {Learning}}, ser. {ICML}'10.\hskip 1em plus 0.5em minus 0.4em\relax Omnipress, 2010.

\bibitem{kingma_adam_2015}
D.~P. Kingma and J.~Ba, ``\href{https://arxiv.org/abs/1412.6980}{Adam: {A} {Method} for {Stochastic} {Optimization}},'' in \emph{3rd {International} {Conference} on {Learning} {Representations}, {ICLR} 2015 {Conference} {Track} {Proceedings}}, 2015.

\bibitem{ferri_comparative_1994}
F.~Ferri, P.~Pudil, M.~Hatef, and J.~Kittler, ``\BIBforeignlanguage{en}{\href{https://linkinghub.elsevier.com/retrieve/pii/B9780444818928500407}{Comparative study of techniques for large-scale feature selection}},'' in \emph{\BIBforeignlanguage{en}{Machine {Intelligence} and {Pattern} {Recognition}}}.\hskip 1em plus 0.5em minus 0.4em\relax Elsevier, 1994, vol.~16.

\bibitem{kendon1981geography}
A.~Kendon, ``Geography of gesture,'' \emph{Semiotica}, vol.~37, no. 1/2, pp. 129--163, 1981.

\end{thebibliography}

\end{document}